\documentclass[12pt]{article}
\usepackage[english]{babel}
\usepackage{graphicx,color}
\usepackage{framed}
\usepackage[normalem]{ulem}
\usepackage{amsmath}
\usepackage{amsthm}
\usepackage{amssymb}
\usepackage{amsfonts}
\usepackage{enumerate}
\usepackage{algorithm}
\usepackage{algpseudocode}
\usepackage[top=1 in,bottom=1in, left=0.8 in, right=0.8 in]{geometry}
\usepackage{bm}
\usepackage{natbib}
\usepackage{multirow}
\usepackage{makecell}
\usepackage{comment}

\theoremstyle{definition}

\usepackage{xcolor}

\usepackage{setspace}

\AtBeginDocument{%
  \doublespacing
  \setlength{\textheight}{\dimexpr 27\baselineskip+\topskip\relax}
}

\usepackage{hyperref}
\hypersetup{
    colorlinks=true,
    linkcolor=black,
    filecolor=magenta,      
    urlcolor=cyan,
    citecolor = blue,
    pdftitle={Overleaf Example},
    pdfpagemode=FullScreen,
    }

\usepackage {xr}
\makeatletter
\newcommand*{\addFileDependency}[1]{
  \typeout{(#1)}
  \@addtofilelist{#1}
  \IfFileExists{#1}{}{\typeout{No file #1.}}
}
\makeatother

\newcommand*{\myexternaldocument}[1]{
    \externaldocument{#1}
    \addFileDependency{main.tex}
}

\myexternaldocument{SM}
\usepackage{etoolbox}
\AtBeginEnvironment{thebibliography}{
    \small
    \singlespacing
}

\usepackage{titlesec}
\titleformat{\section}
  {\Large\bfseries}
  {\thesection}
  {0.5em}
  {}

\titlespacing{\section}
  {0pt}
  {5pt}
  {0pt}

\titleformat{\subsection}
  {\bfseries\large}
  {\thesubsection}
  {0.5em}
  {}

\titlespacing{\subsection}
  {0pt}
  {0pt}
  {0pt}

\titlespacing*{\paragraph}
  {0pt}
  {0pt}
  {1em}

\AtBeginEnvironment{equation}{%
  \setlength{\abovedisplayskip}{3pt}
  \setlength{\belowdisplayskip}{3pt}
}
\AtBeginEnvironment{equation*}{%
  \setlength{\abovedisplayskip}{3pt}
  \setlength{\belowdisplayskip}{3pt}
}

\AtBeginEnvironment{eqnarray}{%
  \setlength{\abovedisplayskip}{3pt}
  \setlength{\belowdisplayskip}{3pt}
}
\AtBeginEnvironment{eqnarray*}{%
  \setlength{\abovedisplayskip}{3pt}
  \setlength{\belowdisplayskip}{3pt}
}
\AtBeginEnvironment{equation*}{%
  \setlength{\abovedisplayskip}{3pt}
  \setlength{\belowdisplayskip}{3pt}
}

\title{Comments on “Deep Fr\'echet Regression”}
\date{}
\author{Jianbin Tan\thanks{Department of Biostatistics \& Bioinformatics, Duke University} ~ and ~ Anru R. Zhang\thanks{Department of Biostatistics \& Bioinformatics and Department of Computer Science, Duke University}}

\begin{document}

\maketitle

% \begin{abstract}
% \singlespacing

% \end{abstract}

% \noindent%
% {\it Keywords:} 

\begin{sloppypar}

Deep Fr\'echet regression \citep{Iao03072025} provides an elegant solution to nonlinear regression
with metric-space-valued responses $Y$ and potentially high-dimensional
Euclidean predictors $X$. A distinctive feature of the proposed approach is
its multi-step construction. 
It first maps the observed response objects $Y$
to a low-dimensional Euclidean representation $Z$ using manifold learning,
then uses deep neural networks to learn the relationship between the covariates
$X$ and the Euclidean coordinates $Z$, and finally employs local Fr\'echet regression to reverse-map the predicted
coordinates back to the original response space.

This construction reflects a fundamental difficulty of the problem: the
regression procedure must repeatedly bridge Euclidean vectors and
non-Euclidean objects. Meanwhile, the manifold-to-Euclidean mapping is
constructed without directly incorporating covariate information, with the
covariates entering only through the intermediate regression from $X$ to $Z$.
We therefore raise the question of whether there may be an
alternative route: rather than constructing covariate-independent mappings
between objects of different types, could one instead reformulate the problem
through a covariate-dependent transport between objects of the same type?

\begin{figure}[h]
    \centering
    \includegraphics[width=0.7\textwidth]{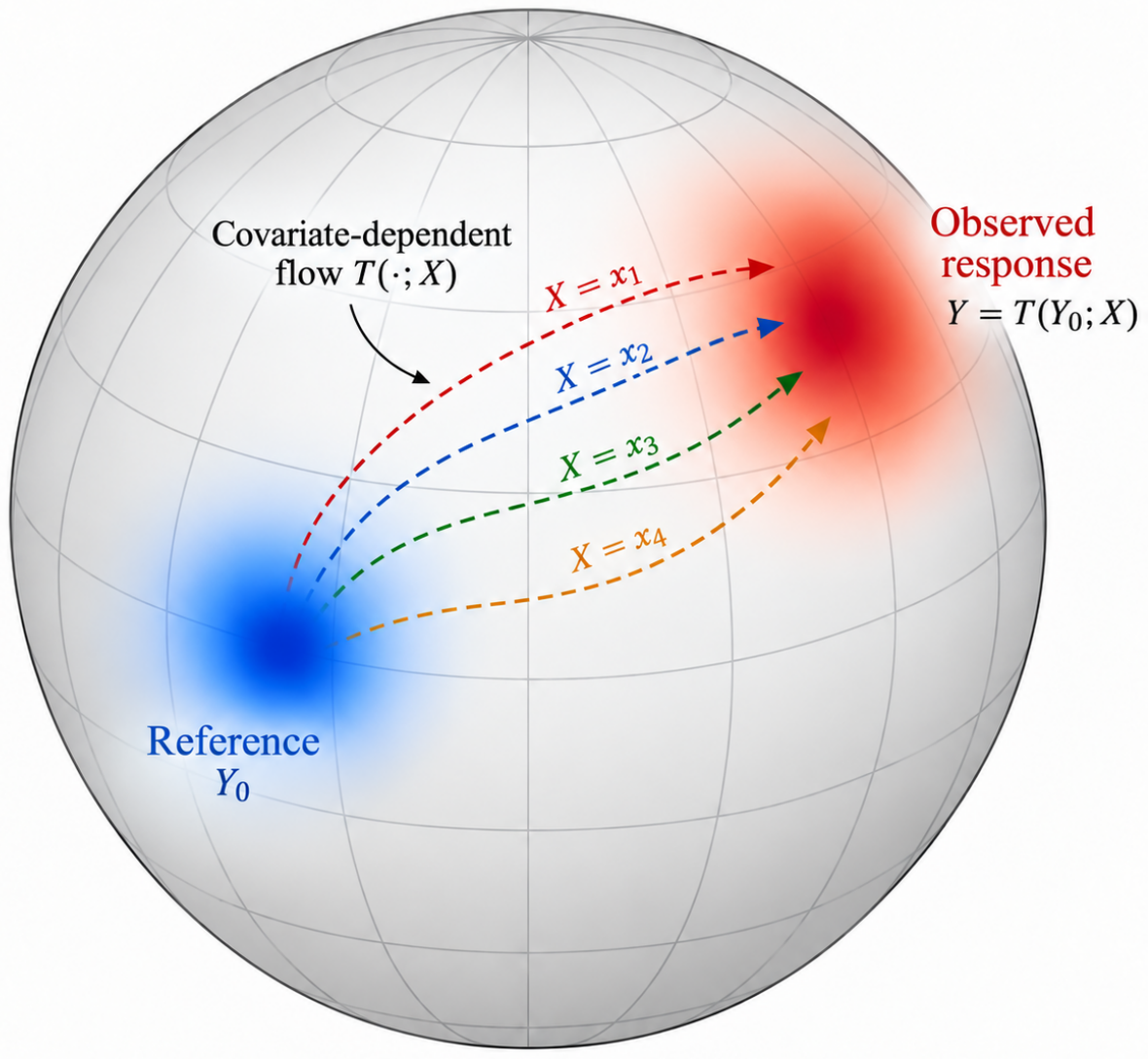}
    \caption{Illustration of a covariate-dependent flow on a non-Euclidean
    response space, transporting a reference distribution toward the conditional
    distribution of the observed response.}
    \label{fig:flow_transport}
\end{figure}

For response spaces with known geometric structure, one possible solution is to
start from a reference distribution on the response space that is easy to
sample from, and then learn a covariate-dependent transport to the conditional
distribution of the observed responses. 
This perspective is closely related to {\it flow matching}, which aims to learn
continuous transport/flow between probability distributions over the same
ambient space
\citep{liu2022flow,lipman2023flow,tong2023improving}.
See Figure~\ref{fig:flow_transport} for an illustration of covariate-driven
transport on a non-Euclidean response space.

Such a formulation may offer several advantages. First, a continuous transport
within the same response space is naturally reversible under standard
regularity conditions. It therefore does not require separate manifold-learning and
reverse-mapping procedures, potentially reducing error propagation across
multiple estimated components.
Second, the
geometry of the response space and the covariate information could jointly
determine the transport, rather than being incorporated through a
covariate-independent representation followed by regression. Third, it
naturally targets the full conditional distribution $P(Y\mid X)$ rather than
focusing only on a summary such as the conditional Fr\'echet mean.

Although this perspective is still preliminary, related flow-matching ideas
have been explored on Riemannian manifolds and in other non-Euclidean
settings
\citep{chen2024flow,haviv2024wasserstein,kapusniak2024metric,qin2024defog,tan2025smooth,wyrwal2026topological}.
These developments suggest that flow modeling in non-Euclidean spaces may
offer a complementary perspective for structured-response regression. We hope
that this viewpoint may help motivate further exploration at the intersection
of deep Fr\'echet regression and generative transport methods.

\section*{Acknowledgement}
This research was supported in part by NIH Grant R01HL169347.

\section*{Conflict of Interest}

The authors report there are no competing interests to declare.

\bibliographystyle{chicago}
\bibliography{refbib}

\end{sloppypar}

\end{document}